\documentclass[a4paper, 10pt, conference]{ieeeconf}      

\IEEEoverridecommandlockouts                              

\usepackage{graphics} 
\usepackage{epsfig} 
\usepackage{times} 
\usepackage{amsmath} 
\usepackage{amssymb}  
\usepackage{kotex}

\title{\LARGE \bf
Model Comparison for Fast Domain Adaptation \\in Table Service Scenario*
}

\author{Woo-han Yun$^{1}$, Minsu Jang$^{1}$, and Jaehong Kim$^{1}$
\thanks{*This work was supported by Institute for Information \& communications Technology Promotion(IITP) grant funded by the Korea government(MSIP) (No.2020-0-00842, Development of Cloud Robot Intelligence for Continual Adaptation to User Reactions in Real Service Environments)}
\thanks{$^{1}$Woo-han Yun, Minsu Jang, and Jaehong Kim are with the Human Robot Interaction Research Section, Mobility Robot Research Division, 
        Electronics and Telecommunications Research Institute, 218 Gajeong-ro, Yuseong-gu, Daejeon, the Republic of Korea
        {\tt\small yochin@etri.re.kr, minsu@etri.re.kr, jhkim504@etri.re.kr}}%
}

\begin{document}

\maketitle
\thispagestyle{empty}
\pagestyle{empty}

\begin{abstract}
In restaurants, many aspects of customer service, such as greeting customers, taking orders, and processing payments, are automated.
Due to the various cuisines, required services, and different standards of each restaurant, one challenging part of making the entire automated process is inspecting and providing appropriate services at the table during a meal.
In this paper, we demonstrate an approach for automatically checking and providing services at the table.
We initially construct a base model to recognize common information to comprehend the context of the table, such as object category, remaining food quantity, and meal progress status.
After that, we add a service recognition classifier and retrain the model using a small amount of local restaurant data.
We gathered data capturing the restaurant table during the meal in order to find a suitable service recognition classifier.
With different inputs, combinations, time series, and data choices, we carried out a variety of tests.
Through these tests, we discovered that the model with few significant data points and trainable parameters is more crucial in the case of sparse and redundant retraining data. 

\end{abstract}

\section{INTRODUCTION}

With the recent development of deep neural networks, many of the tasks that were previously done by humans have been automated such as harvesting fruit and vegetables \cite{ur_harvest} and autonomous driving \cite{ur_driving}\cite{ur_driver}.
One possible task that can be automated is customer service in restaurants.
In customer services, there are welcoming guests, checking table conditions and guiding them to their seats, taking orders, checking and providing necessary services at the table, and processing payments.
Many of these services could be replaced by tabletop robots (including tablets) and serving robots, but there is no service for checking and providing necessary services that is taking most of the time.
The variety of foods, services, and criteria for providing services by each restaurant make it difficult to automate these services.

In this paper, we explain how we tried to automate these checking and providing necessary services at the table, called a table service, and what we learned in this experiment.
Overall, we leveraged deep learning methods to build a process.
Deep learning requires massive data and a complex model to achieve good performance.
Because it is difficult to get large amounts of data at individual restaurants, we first build a base model to recognize common cues and retrain the model to individual restaurants using transfer learning techniques.
The common cues recognized in the base model are the location and category of food and things, the remained amount of foods, and the progress status of meals.
We believed that this information is common and essential to understanding the context of the restaurant in most restaurants.
The next assumption is that we could get a handful of data, that is images and labels, at each restaurant.
Using this local restaurant data, we retrain the network for recognizing table service alarms.
We considered four cases of table service alarms, food refill, garbage collection, dessert provided, and check lost.
In summary, we build a global recognition model for three cues for all restaurants and retrain the model for recognizing four cases of table service alarms for individual restaurants, separately.

\section{Related Work}
\subsection{Object Detection}
As a base model, we consider object detection with two heads for leftover food and the progress of meals.
Object detection is a task to find the location and category of objects.
The architecture for object detection is made up of a backbone network for feature extraction and a task network for localization and classification.
A backbone network is normally borrowed from a massively pre-trained model for classification tasks because a strong feature extractor requires a massive training dataset.
Depending on the presence of generic object proposals, the detection network is categorized into two types: one-staged and two-staged architecture.
In one-staged architectures \cite{Redmon2015a-YOLO}\cite{Liu2015-ParseNet}, the location and category of objects are directly estimated from whole image features.
On the other hand, the generic object proposal (including only objectness and location) is estimated at the first stage, then the category and detailed location of objects is estimated at the second stage in two-staged architecture \cite{Girshick2015fastrcnn}\cite{Ren2017-fasterrcnnRe}
Recently, DETR (DEtection with TRansformer) \cite{DETR} and its variants \cite{DeformDETR}\cite{Dai2012_DynamicDETR} showed a detector without hand-designed components such as NMS (Non-maxima suppression) using transformer architecture.

\subsection{Active Learning}
The trained model is typically degraded in situations when training and testing differ.
Choosing relevant data from an unlabeled test set and retraining the network are steps in the process of making the model more reliable and reducing the cost of labeling.
In active learning, the meaningful data is selected based on the two criteria, diversity \cite{al_coreset} and uncertainty \cite{al_uncertainty}.
In the uncertainty-based approach, the data that is uncertain in the output of the current model is selected.
On the other hand, the data representing the entire data space is selected in a diversity-based approach.

\section{Approach}

\subsection{Table Information Recognition}
\subsubsection{Model}
\begin{figure}[t]
  \centering
  \includegraphics[width=0.5\textwidth]{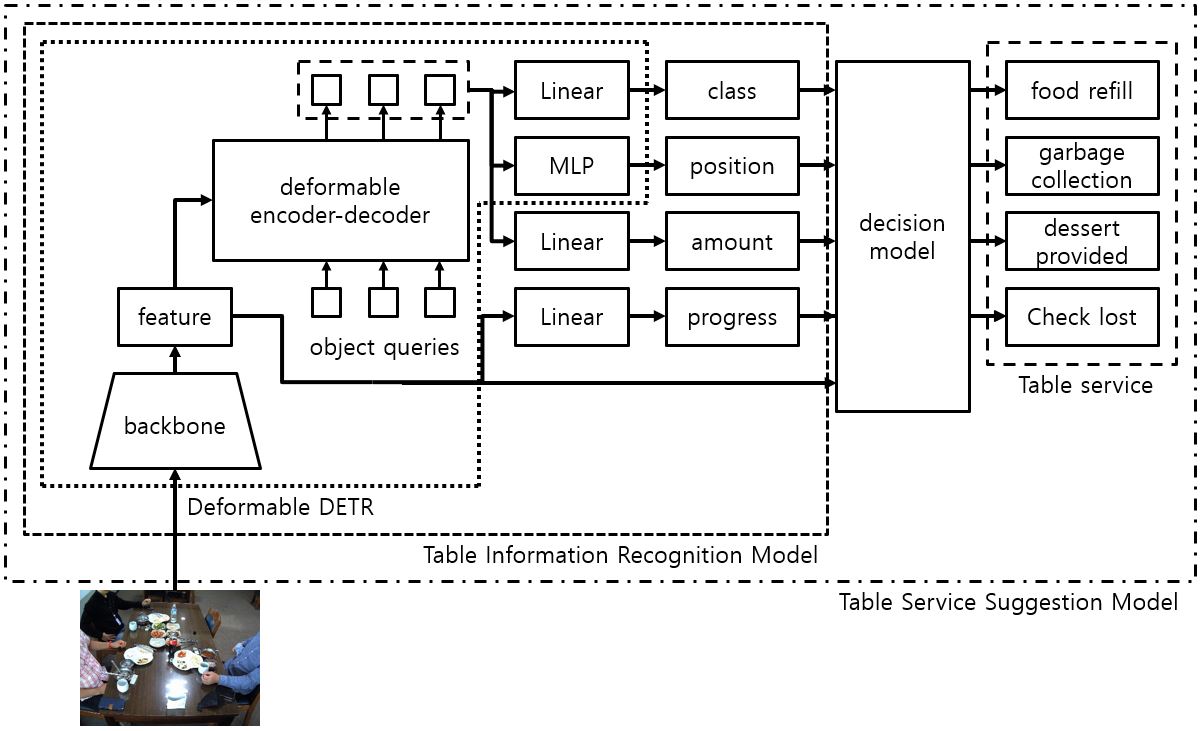}
  \caption{Table Information Recognition and Service Suggestion Model}
  \label{fig:tableServiceModel}
\end{figure}

Fig.~\ref{fig:tableServiceModel} shows a Deformable DETR with two more heads for recognizing the amount of food and the progress status of meals.
Deformable DETR \cite{DeformDETR} is a detection architecture composed of a CNN-based backbone and a transformer-based encoder-decoder.
It is based on DETR \cite{DETR} which removed hand-designed components such as NMS (Non-maxima suppression) and allows faster training by making the attention module refer only to sampled partial regions.
In our work, we additionally added one head at region-based features to recognize the amount of food and the other head at backbone features to estimate the progress status of the meals.
For estimating the amount of remaining food, we used a classification model instead of a regression model because of a better recognition result.

\subsubsection{Dataset and Training}
\begin{figure}[t]
  \centering
  \includegraphics[width=0.48\textwidth]{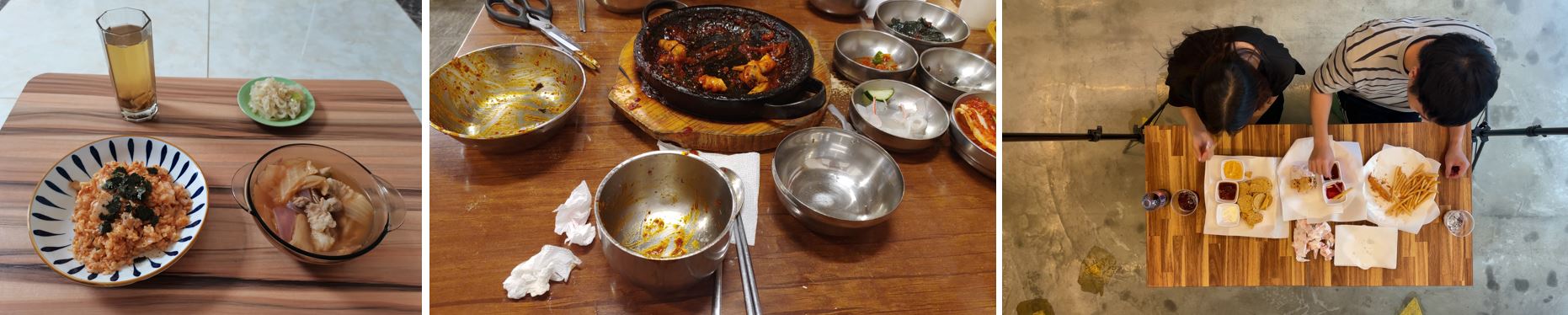}
  \caption{Example images from three datasets, CloudStatus, CloudTableThings, and CloudMeals.}
  \label{fig:db_basemodel}
\end{figure}

Since there is no dataset including all three types of label information, we train the model by interleaving three datasets having different types of labels.
Three datasets we used in training the table information recognition model are in Table~\ref{tab:4dbs} and Figure~\ref{fig:db_basemodel}.
We divided the dataset at a ratio of 9 and 1 for training and evaluation.

\begin{table}[h]
\caption{Three types of datasets for training table information recognition model. p, d, and a means the progress status of meals, detection category, and amount of foods, respectively.}
\label{tab:4dbs}
\begin{center}
\begin{tabular}{|c||c|c|c|}
\hline
info.           & TableStatus\cite{TableStatus}    & TableThings     & CloudMeals       \\
\hline
\# of images    & 10,064                & 4,000           & 13,313           \\
\hline
label types     & p, d, a               & d               & d, a             \\
\hline
\end{tabular}
\end{center}
\end{table}

The Table Information Recognition model has a total of four losses for optimizing four outputs: object category $loss_{c}$ and its bounding box position $loss_{bb}$, amount of remained food $loss_{a}$, and the progress status of meals $loss_{prog}$ in eq.(1).

$$
loss = loss_{c} + loss_{bb} + loss_{a} + loss_{prog} \eqno{(1)}
$$

The model is trained with a learning rate 2e-4 for 40 epochs and 2e-5 for more 10 epochs.
We followed the training details in \cite{DeformDETR}.

\subsection{Table Service Suggestion Model}
\subsubsection{Dataset and Training}
\begin{figure}[t]
  \centering
  \includegraphics[width=0.48\textwidth]{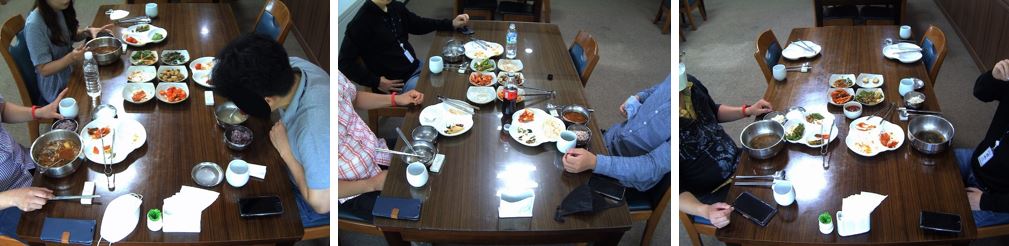}
  \caption{Example images for table service suggestion model.}
  \label{fig:db_servicemodel}
\end{figure}

As a local restaurant dataset, we collected images and tagged labels at the cafeteria in the company, ETRI. 
A total of nine videos were recorded, two videos were used for retraining and seven were used for evaluation.
For retraining, we selected two videos containing all four types of service alarm cases.
We sampled and used 1 fps for training and evaluation.
The number of images and labels are summarized in Table~\ref{tab:3dbs} and Figure~\ref{fig:db_servicemodel}.

The model is trained with a learning rate 1e-2 for 40 epochs, 1e-3 for 40 epochs, and 1e-5 for more 20 epochs.

\begin{table}[h]
\caption{Dataset for table service suggestion model}
\label{tab:3dbs}
\begin{center}
\begin{tabular}{|c||c||c|c|c|c|}
\hline
dataset    & \# of images   & refill    & trash    & dessert  & lost      \\
\hline
train      & 2,440          & 2,301      & 627      & 572      & 99        \\
test       & 9,848          & 6,622      & 3,613     & 2,319     & 101       \\
\hline
\end{tabular}
\end{center}
\end{table}

\subsubsection{Task Definition}
We approached the table service suggestion as a table service recognition problem.
Given an input image, the model tried to find the proper table service from four service categories.
As a baseline model, we used a simple classification model that is based on a CNN-based backbone and one fully-connected layer after the average pooling layer.
For a loss, the cross-entropy loss was used.
Because the dataset is highly imbalanced as in \ref{tab:3dbs}, we used two metrics for comparing models, f1 score and ROC AUC.


\subsection{Retraining}
\subsubsection{Frozen the backbone}
We first tested whether the backbone would be frozen or trainable.
In our test, we always got a higher recognition rate with a frozen backbone rather than a trainable backbone.
From this observation, we froze the backbone in the following experiments.

\subsubsection{Feature Selection}
We experimented and compared the performance of models under various input features and combinations of features.
The available inputs for the table service classification are images-based features from the CNN-based backbone and encoder and region-based features from the decoder, and the three outputs (object category, food amount, and progress of meals) of table information models with time spent eating in seconds.
Because the three outputs are estimated from the model, the outputs might contain errors.
The results are in Table~\ref{tab:res_input_avgp}. 
The baseline model using the CNN-based backbone feature showed the best result and the others were in the order of those using encoder features, decoder features, and recognition outputs.
The image-based features (backbone and encoder) showed better performance than the region-based features (decoder and recognition).

Next, we conducted a similar experiment by using a simple attention consisting of linear and softmax layers instead of the average pooling.
With simple attention, each pixel of image-based features and each region of region-based features has its attention weight and is merged with this weight.
The result is in Table~\ref{tab:res_input_softattn}.
From the average pooling to simple attention in Table~\ref{tab:res_input_avgp} and Table~\ref{tab:res_input_softattn}, the result using image-based features (backbone and encoder) is worse and those using region-based features (decoder and recognition) are better.

\begin{table}[t]
\caption{Result comparison by various inputs with average pooling.}
\label{tab:res_input_avgp}
\begin{center}
\begin{tabular}{|c||c||c|c|}
\hline
input         & f1 score   & ROC       \\
\hline
backbone      & 59.25      & 88.08             \\
encoder       & 53.17      & 78.86             \\
decoder       & 38.92      & 78.50             \\
table info.   & 35.45      & 68.67             \\
\hline
\end{tabular}
\end{center}
\end{table}

\begin{table}[t]
\caption{Result comparison by various inputs with simple attention.}
\label{tab:res_input_softattn}
\begin{center}
\begin{tabular}{|c||c||c|c|}
\hline
input         & f1 score   & ROC               \\
\hline
backbone      & 39.58      & 72.42             \\
encoder       & 40.73      & 81.58             \\
decoder       & 47.41      & 78.07             \\
table info.   & 46.74      & 86.55             \\
\hline
\end{tabular}
\end{center}
\end{table}

As a next step, we aggregate more features based on the backbone with average pooling showing the best results.
We select top-4 features as candidate features.
The comparison is in Table~\ref{tab:res_input_combi}.
Interestingly, the best combination was the backbone with average pooling and table information with simple attention those are the most low-level and high-level features.
We used these two features in the following experiments.

\begin{table}[h]
\caption{Result comparison by combining features. The CNN-based backbone with average pooling is not in the table but was used by default. }
\label{tab:res_input_combi}
\begin{center}
\begin{tabular}{|c|c||c||c|c|}
\hline
input       & aggre.        & f1 score   & ROC       \\
\hline
encoder     & avg.pool      & 57.01      & 88.58             \\
encoder     & soft attn.    & 47.07      & 86.78             \\
decoder     & soft attn.    & 59.21      & 87.09             \\
table info. & soft attn.    & 65.74      & 92.23             \\
\hline
\end{tabular}
\end{center}
\end{table}

\subsubsection{Image to sequence}
Next, we tested with a video-based approach instead of an image-based approach.
As a sequence, we used five frames of five seconds.
We considered simple attention, max pooling, and average pooling for modeling temporal sequence.
In Table~\ref{tab:res_video}, all models using sequence data are worse than the image-based model.

\begin{table}[h]
\caption{Result comparison by various inputs with simple attention }
\label{tab:res_video}
\begin{center}
\begin{tabular}{|c||c||c|c|}
\hline
temporal aggre.  & f1 score   & ROC       \\
\hline
image            & 65.74      & 92.23             \\
max pooling      & 62.30      & 88.35             \\
avg pooling      & 63.26      & 89.56             \\
Soft Attn.        & 64.43      & 87.42             \\
\hline
\end{tabular}
\end{center}
\end{table}

\subsubsection{Full dataset or partial dataset}
It is possible that there will not be enough money or time to collect a variety of data when applying the model to local restaurants.
In this case, choosing more significant data with an active learning technique may be an option.
We experimented with two representative active learning approaches, uncertainty-based \cite{al_uncertainty} and diversity-based \cite{al_coreset} methods and a random selection method.
In this experiment, the CNN-based backbone with average pooling served as the baseline model.
The results are in Table~\ref{tab:res_al}.
It is interesting to note that all findings with partial datasets outperform those with full datasets in terms of the f1 score.
With the exception of uncertainty-based methods, partial datasets were more helpful in ROC than whole datasets.
By a wide margin, the diversity-based selection strategy was the best on both metrics.
When the retraining dataset is fairly little and redundant, picking and using a few important data points is more helpful for training. 

\begin{table}[h]
\caption{Result comparison by active data selection }
\label{tab:res_al}
\begin{center}
\begin{tabular}{|c||c|c|}
\hline
selection method & f1 score   & ROC       \\
\hline
all              & 59.25      & 87.09             \\
random           & 61.13      & 87.22             \\
uncertainty      & 71.36      & 85.21             \\
diversity        & 68.92      & 88.04             \\
\hline
\end{tabular}
\end{center}
\end{table}


\subsection{Qualitative Results}
\begin{figure}[t]
  \centering
  \includegraphics[width=0.4\textwidth]{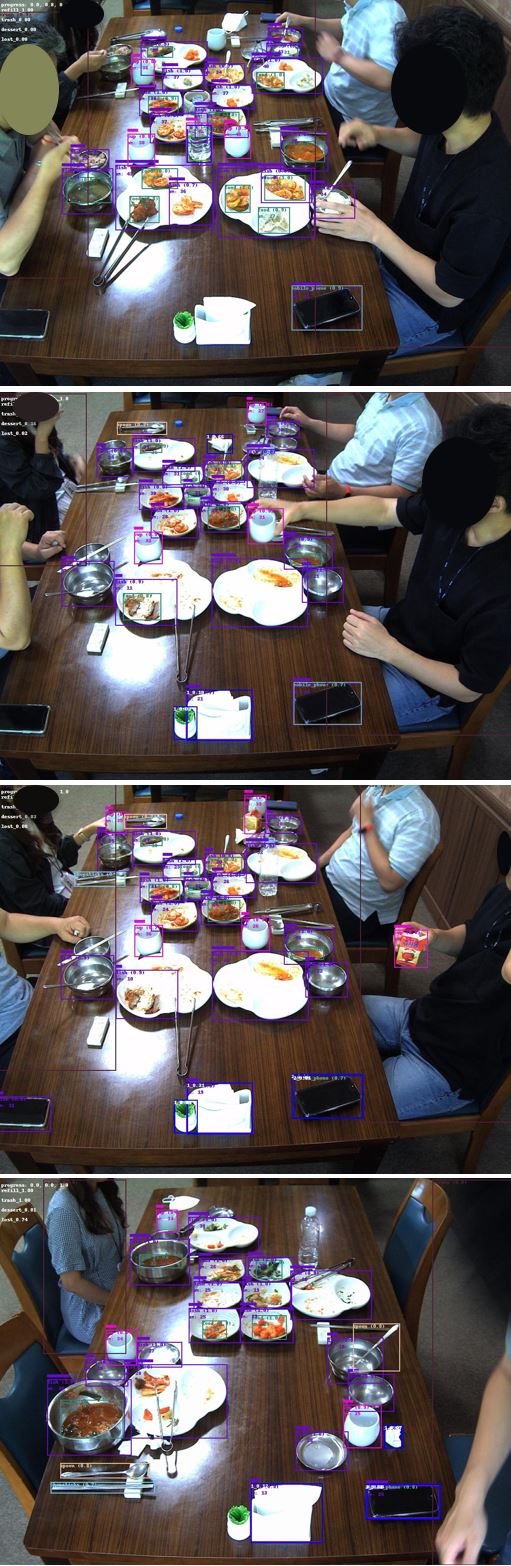}
  \caption{Result images on four service cases. From top to bottom, each result shows the service cases of food refill, garbage collection, providing dessert, and finding lost, respectively.}
  \label{fig:result}
\end{figure}

In Figure~\ref{fig:result}, the example results of four cases are illustrated.
From top to bottom, each result shows the recognized service cases of food refill, garbage collection, providing dessert, and finding lost, respectively.
In the first image, a food refill occurred when some side dishes were almost empty.
The garbage collection request was recognized when there was trash on the top of the table in the second image.
Third, providing dessert requests are recognized when people stop eating and almost dishes are empty.
In the final image, if there is a personal item but no one is present, the lost item alarm is recognized.

In evaluation, our system frequently makes mistakes.
For instance, an improper refill request alarm is produced when a water bottle is detected but its contents are not recognized.
Additionally, detecting a paper tissue in the tissue case at the bottom of the image sets off an erratic alarm, a garbage collection alert.
It will be necessary to reduce these errors of recognition by extra training or adopting constraints.

\section{CONCLUSIONS}
In this paper, we demonstrate our experiment for automatic checking and providing necessary service at the table, called a table service.
To do this, we build a base model for identifying table information and retrain another classifier with decision model for recognizing four table services.
We evaluated the model using a variety of input features, combinations, and sequences in order to find the optimal model. 
We also experimented with partial meaningful dataset usage with active learning.
We conclude from these tests that blending low-level and high-level features is the optimum strategy, even when less trainable parameters are useful in the absence of sufficient data.
Additionally, fewer, useful data points are worth more than many, redundant data points.
We intend to develop more reliable models and use them at actual restaurant demonstration sites in light of this observation. 

\addtolength{\textheight}{-12cm}   



\bibliographystyle{IEEEtran}
\bibliography{library.bib}

\begin{thebibliography}{10}
\providecommand{\url}[1]{#1}
\csname url@rmstyle\endcsname
\providecommand{\newblock}{\relax}
\providecommand{\bibinfo}[2]{#2}
\providecommand\BIBentrySTDinterwordspacing{\spaceskip=0pt\relax}
\providecommand\BIBentryALTinterwordstretchfactor{4}
\providecommand\BIBentryALTinterwordspacing{\spaceskip=\fontdimen2\font plus
\BIBentryALTinterwordstretchfactor\fontdimen3\font minus \fontdimen4\font\relax}
\providecommand\BIBforeignlanguage[2]{{%
\expandafter\ifx\csname l@#1\endcsname\relax
\typeout{** WARNING: IEEEtran.bst: No hyphenation pattern has been}%
\typeout{** loaded for the language `#1'. Using the pattern for}%
\typeout{** the default language instead.}%
\else
\language=\csname l@#1\endcsname
\fi
#2}}

\bibitem{ur_harvest}
J.~Seol, S.~Lee, and H.~I. Son, ``A review of end-effector for fruit and vegetable harvesting robot,'' in \emph{Journal of Korea Robotics Society}, 2020.

\bibitem{ur_driving}
S.~Hossain and D.-J. Lee, ``Autonomous-driving vehicle learning environments using unity real-time engine and end-to-end cnn approach,'' in \emph{Journal of Korea Robotics Society}, 2019.

\bibitem{ur_driver}
X.~Miao, H.-S. Lee, and B.-Y. Kang, ``Development of driver's safety/danger status cognitive assistance system based on deep learning,'' in \emph{Journal of Korea Robotics Society}, 2018.

\bibitem{Redmon2015a-YOLO}
J.~Redmon, S.~Divvala, R.~Girshick, and A.~Farhadi, ``{You Only Look Once: Unified, Real-Time Object Detection},'' in \emph{CVPR}, 2016.

\bibitem{Liu2015-ParseNet}
\BIBentryALTinterwordspacing
W.~Liu, A.~Rabinovich, and A.~C. Berg, ``{ParseNet: Looking Wider to See Better},'' 2015. [Online]. Available: \url{http://arxiv.org/abs/1506.04579}
\BIBentrySTDinterwordspacing

\bibitem{Girshick2015fastrcnn}
R.~Girshick, ``{Fast R-CNN},'' in \emph{ICCV}, 2015.

\bibitem{Ren2017-fasterrcnnRe}
S.~Ren, K.~He, R.~Girshick, and J.~Sun, ``{Faster R-CNN: Towards Real-Time Object Detection with Region Proposal Networks},'' \emph{IEEE PAMI}, 2017.

\bibitem{DETR}
N.~Carion, F.~Massa, G.~Synnaeve, N.~Usunier, A.~Kirillov, and S.~Zagoruyko, ``End-to-end object detection with transformers,'' in \emph{ECCV}, 2020.

\bibitem{DeformDETR}
X.~Zhu, W.~Su, L.~Lu, B.~Li, X.~Wang, and J.~Dai, ``Deformable {DETR:} deformable transformers for end-to-end object detection,'' in \emph{ICLR}, 2021.

\bibitem{Dai2012_DynamicDETR}
X.~Dai, Y.~Chen, J.~Yang, P.~Zhang, L.~Yuan, and L.~Zhang, ``Dynamic detr: End-to-end object detection with dynamic attention,'' in \emph{Proceedings of the IEEE/CVF International Conference on Computer Vision (ICCV)}, October 2021, pp. 2988--2997.

\bibitem{al_coreset}
O.~Sener and S.~Savarese, ``Active learning for convolutional neural networks: A core-set approach,'' in \emph{ICLR}, 2018.

\bibitem{al_uncertainty}
A.~J. Joshi, F.~Porikli, and N.~Papanikolopoulos, ``Multi-class active learning for image classification,'' in \emph{CVPR}, 2009.

\bibitem{TableStatus}
W.~han Yun, M.~Jang, and J.~Kim, ``Performance improvement of food detection using domain adaptation,'' in \emph{Summer Annual Conference of IEIE}, 2022.

\end{thebibliography}

\end{document}